\documentclass[runningheads]{llncs}
\usepackage{graphicx}
\usepackage{amsmath,amssymb} 
\usepackage{color}
\usepackage{subfig}
\def\bm#1{\boldsymbol{#1}}
\begin{document}
	\title{Deriving Neural Network Architectures using Precision Learning: Parallel-to-fan beam Conversion} 
	
	\titlerunning{Deriving Neural Network Architectures using Precision Learning}
	%
	\author{Christopher Syben\inst{1,2} \and Bernhard Stimpel\inst{1,2} \and Jonathan Lommen\inst{1,2} \and Tobias W\"urfl\inst{1} \and Arnd D\"orfler\inst{2} \and Andreas Maier\inst{1}}
	%
	\authorrunning{Syben et al.}
	%
	
	\institute{ Pattern Recognition Lab, Department of Computer Science,
		Friedrich-Alexander-Universit\"at Erlangen-N\"urnberg, Germany \and
		Department of Neuroradiology, Universit\"atsklinikum Erlangen,
		Friedrich-Alexander-Universit\"at Erlangen-N\"urnberg, Germany}

	\maketitle              
	\begin{abstract}
		In this paper, we derive a neural network architecture based on an analytical formulation of the parallel-to-fan beam conversion problem following the concept of precision learning. The network allows to learn the unknown operators in this conversion in a data-driven manner avoiding interpolation and potential loss of resolution. Integration of known operators results in a small number of trainable parameters that can be estimated from synthetic data only.  
		The concept is evaluated in the context of Hybrid MRI/X-ray imaging where transformation of the parallel-beam MRI projections to fan-beam X-ray projections is required. The proposed method is compared to a traditional rebinning method.
		The results demonstrate that the proposed method is superior to ray-by-ray interpolation and is able to deliver sharper images using the same amount of parallel-beam input projections which is crucial for interventional applications.
		We believe that this approach forms a basis for further work uniting deep learning, signal processing, physics, and traditional pattern recognition.
		\keywords{Machine Learning \and Precision Learning \and Hybrid MRI/X-ray imaging }
	\end{abstract}
	\section{Introduction}
	\label{sec:introduction}
	Deep learning is a game-changer in many perceptual tasks ranging from image classification over segmentation to localization \cite{christlein2017tutorial}. A major disadvantage of perceptual problems is that no prior knowledge on how the classes and labels are obtained is available. As such a large body of literature exists that investigates different network topologies for different applications. As result, we managed to replace \textit{hand-crafted features} with \textit{hand-crafted networks}.
	
	Recently, these techniques also emerge to other fields in signal processing. One of them is medical image reconstruction in which surprising results have been obtained \cite{zhu2018image,yixing2018deep}.
	For signal processing, however, we do have prior knowledge available that can be reused in the network design. The use of these prior operators reduces the number of unknowns of the network, therewith the amount of required training samples, and the maximal training error bounds \cite{precision_learning}. Up to now, this \textit{precision learning} approach was only used to augment networks with prior knowledge and or to add more flexibility into existing algorithms \cite{wuerflM,wuerflJ,filter_learning,vesselnet}. In this paper, we want to extend this approach even further: we demonstrate that we can derive a mathematical model to tackle a problem under consideration and use deep learning to formulate different hypothesis on efficient solution schemes that are then found as the point of optimality of a deep learning training process.
	
	In particular, we aim in this paper at an efficient convolution-based solution for parallel-to-fan-beam conversion. Up to now, such an efficient algorithm was unknown and the state-of-the-art to address this problem is rebinning of rays that is inherently connected to interpolation and a loss of resolution. 
	
	The problem at hand is not only interesting in terms of algorithmic development, it also has an immediate application. Novel hybrid medical scanners will be able to combine Computed Tomography (CT) and Magnetic Resonance Imaging (MRI) in a single device for interventional applications \cite{fahrig,wang_hybrid}. While CT offers high spatial and temporal resolution, MRI allows for the visualization of soft-tissue contrast, vessels without the use of contrast agent, and there is no need for harmful ionizing radiation. 
	
	However, acquisition on MR devices is slow compared to CT. Flat-panel detectors allow image-guided interventions using fluoroscopic projection images that can be acquired at high frame-rates with up to 30 frames per second. This is a challenging time constraint for  MRI. Recent developments indicate that MRI is also able to perform projection imaging at acceptable frame rates \cite{mr_projection}. Yet the two modalities are inherently incompatible, as MRI typically operates in a parallel projection geometry and X-rays emerge from a source point that restricts them to fan- and cone-beam geometries.
	
	Recent publications elaborate on the idea of MR/X-ray projection fusion and extend the MR acquisition such that the final MR-projection image shows the same perspective distortion as the X-ray projection \cite{beams_view,igic,mr_projection}. Current approaches, however, rely on rebinning that requires interpolation which inherently reduces the resolution of the generated images. In this paper, we propose to derive an image rebinning method from the classical theory. However, as this would require an expensive inverse of a large matrix, we propose to replace the operation with a highly efficient convolution that is inspired by the classical filtered back-projection solution in CT. Here, we examine two cases for this convolution: a projection-independent and a projection-dependent one.

	\section{Methods}
	In the first section we shortly describe the link between X-ray and MRI projections using rebinning \cite{igic}, afterwards we revisit the discrete form of the reconstruction problem, which is then followed by our proposed problem description. Subsequently the network topology will be derived following the precision learning paradigm. This section is concluded by a description of the training process and the used training data.
	
	\subsection{Linking MRI and X-ray Acqusition}
	The link between the X-ray and MRI acquisition is given by the central slice theorem. This has first been demonstrated by Syben et al. \cite{igic} for simulation data and was later applied for the construction of X-ray projections from MRI measurement data \cite{mr_projection}. Their approach is inspired by the geometric rebinning method which allows the reconstruction of fan-beam data by resampling the fan-beam acquisition to a parallel-beam acquisition. 
	
\begin{figure}[tb]
		\includegraphics[width=\textwidth]{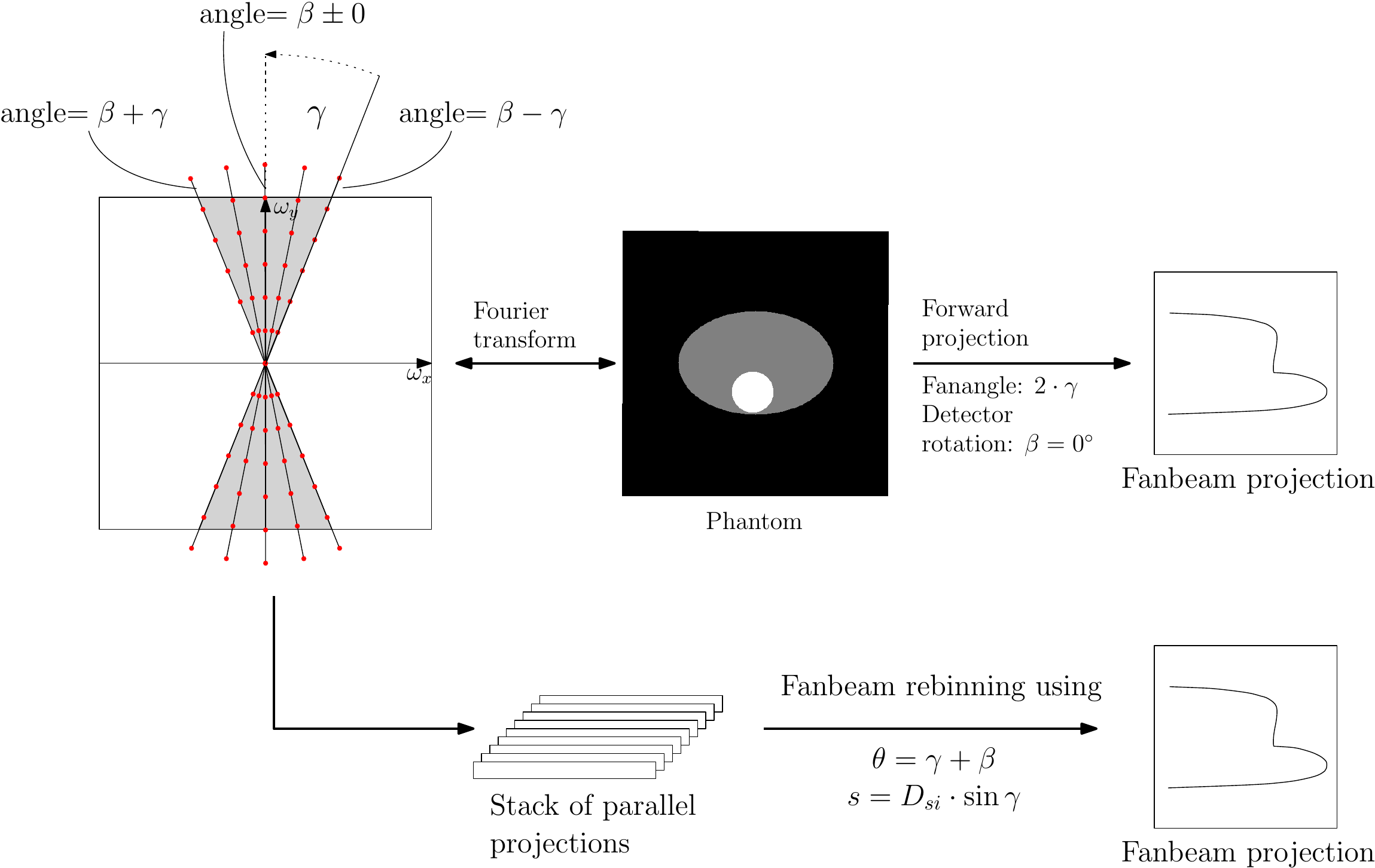}
		\caption{MRI to X-ray link and geometric rebinning method proposed by Syben et al. \cite{igic}.}
		\label{fig:igic_method}
	\end{figure}

	They follow the central slice theorem which states that the Fourier transform of a 1D projection of a 2D object can be found in the 2D Fourier transform of the object along a radial line with the same orientation as the detector. Because the MRI can sample the Fourier transform of the object, parallel projections can be acquired.
	This relationship combined with the geometric rebinning method can be used to convert a set of parallel projections to one fan-beam projection as  shown in Fig \ref{fig:igic_method}.

	In their publication they analyze the sub-sampling capability of this method. In this context,  full sampling means that the MR device acquires one parallel projection for each fan-beam detector pixel. Thus sub-sampling is related to the case where less parallel projections are acquired with respect to the number of fan-beam detector pixels. They show that only few projections are necessary to create the target fan-beam projection with a small error \cite{igic}. Following their geometric rebinning method two steps of interpolation in spatial domain are required: first an interpolation between two projections with different projection angles is carried out followed by an interpolation between the pixels of the parallel-beam projection. 
	\subsection{The Tomographic Reconstruction Problem}
	\label{s:recoProblem}
	The CT imaging procedure from acquiring X-ray projections to the reconstructed object information can be described in discrete linear algebra. The acquisition of the projection images of the object can be described with
	\begin{equation}
	\bm A \bm x = \bm p \enspace,
	\end{equation}
	where $\bm A$ is the system matrix describing the geometry of the imaging system. $\bm x$ is the object itself and $\bm p$ are the projections of $\bm x$ under the described geometry $\bm A$. Correspondingly the reconstruction can be obtained with
	\begin{equation}
	\bm x = \bm A^{-1} \bm p \enspace,
	\end{equation}
	where $\bm A^{-1}$ is the inverse of the system matrix, which can not be inverted since it is a tall matrix. Thus, the reconstruction is conducted using the left-side pseudo inverse which gives the approximation with minimal distance to the inverse in a $\ell$2-norm sense.
	\begin{equation}
	\bm x = \bm A^\top (\bm A \bm A^\top)^{-1} \bm p \label{eq:proj}
	\end{equation}
	where $\bm A^\top$ is the transposed system matrix, which can be algorithmically described as the back-projection operator. For a full scan with 180$^\circ$ of rotation in parallel geometry, the inverse bracket is a filtering step in the Fourier domain and can be described as
	\begin{equation}
	\bm x = \bm A^\top \bm F^{\bm H} \bm K \bm F \bm p
	\end{equation}
	where $\bm F, \bm F^{\bm H}$ is the Fourier and inverse Fourier transform, respectively. $\bm K$ is the so called Ramp filter represented in a diagonal matrix. Together the pseudo inverse this describes the filtered back-projection algorithm in a discrete fashion.
	
	\subsection{Rebinning using Tomographic Reconstruction}
	As shown in \cite{filter_learning}, the discrete description of the reconstruction problem can be used to derive a network topology and to learn the reconstruction filter. In the following, we use this idea to derive an optimization problem to find a filter which can be used to transform several parallel projections to one fan-beam projection.
	A fan-beam projection can be created by
	\begin{equation}
	\bm A_{f} \bm x = \bm p_{f} \enspace,
	\label{eq:fan_proj}
	\end{equation}
	where $\bm A_{f}$ describes the system matrix for a fan-beam projection and $\bm p_{f}$ is the respective projection. The necessary parallel projections which contain the information for the fan-beam projection can be found in the Fourier domain (or K-space of the MRI system) in a wedge region \cite{igic} which is defined by the fan angle of the fan-beam geometry. These parallel projections can be described with
	\begin{equation}
	\bm A_{p} \bm x = \bm p_{p} \enspace,
	\label{eq:par_proj}
	\end{equation}
	where $\bm A_{p}$ is the system matrix generating the projections $\bm p_{p}$ from object $\bm x$ under the parallel-beam geometry. The object $\bm x$ in Eq. \ref{eq:fan_proj} can be substituted by the reconstruction using the inverse of the system matrix and the projections from Eq.~\ref{eq:proj} in Section \ref{s:recoProblem}:
	\begin{equation}
	\bm A_{f} \underbrace{\bm A_{p}^{\top} (\bm A_{p} \bm A_{p}^{\top})^{-1} \bm p_{p}}_{\text{Parallel reconstruction $\bm x$}} = \bm p_{f} \enspace.
	\end{equation}
	In principle, the above equation is hard to solve, as the reconstruction task from this very small set of projections is ill-posed and there is no analytical closed-from solution known. However, we now simply postulate that there exists a projection-independent filter which is a close approximation the above inverse bracket. As in Section~\ref{s:recoProblem}, this allows us to  express the solution as a multiplication with an diagonal filter matrix $\bm K$ in Fourier domain:
	\begin{equation}
	\bm A_{f} \bm A_{p}^{\top}\bm F^{\bm H}\bm K \bm F \bm p_{p} = \hat{\bm p}_{f} \enspace, \label{eq:netequation}
	\end{equation}
	where $\hat{\bm p}_{f}$ is the approximated fan-beam projection under the above stated assumption.
	Now the only unknown operation in above equation is $\bm K$ that can be determined using an objective function:
	\begin{equation}		
	f(\bm K) = \frac{1}{2} \lVert \bm A_{f} \bm A_{p}^{\top} \bm F^{\bm H}\bm K\bm F\bm p_{p} - \bm p_{f} \rVert_2^2 \enspace.		
	\label{eq:objective_function}		
	\end{equation}
	The gradient of function $f$ is with respect to $\bm K$ is	
	\begin{equation}
	\frac{\partial f(\bm K)}{\partial\bm K} = \bm F\bm A_{p}\bm A_{f}^{\top}(\bm A_{f} \bm A_{p}^{\top} \bm F^{\bm H}\bm K\bm F\bm p_{p} - \bm p_{f})(\bm F\bm p_{p})^{\top} \enspace.		
	\label{back_prop}		
	\end{equation}
	Note that this gradient is determined automatically by back-propagation to update the weights of layer $\bm K$, if Eq.\ref{eq:netequation} is implemented by means of a neural network as already observed for a different application in \cite{filter_learning}. Thus, the network topology for a network which learns the transformation from several parallel projections to one fan-beam projection could be derived by the presented approach.	
	\subsection{Network}
\begin{figure}
		\includegraphics[width=\textwidth]{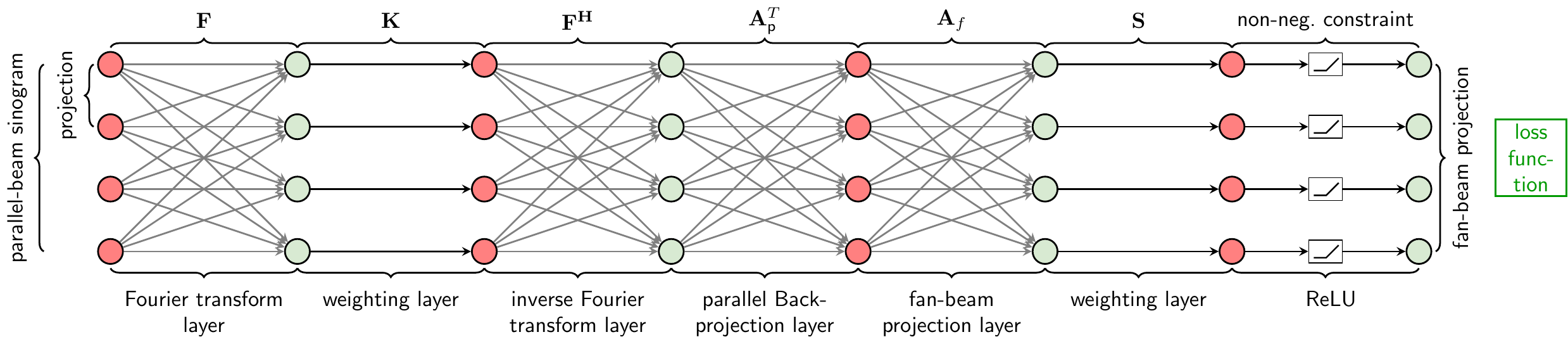}
		\caption{Network based on objective function (Eq. \ref{eq:objective_function}).}
		\label{fig:network}
	\end{figure}

	The network topology can be directly derived from the description of the objective function in Eq. \ref{eq:objective_function} and is shown in Fig. \ref{fig:network}.
	Projectors and back-projectors are scaled to each other in terms of sampling density and number of projections. Since we mix a parallel back-projector with a fan-beam forward projector and aim at different sampling densities we added an additional scaling layer $\bm S$ to the network to compensate accordingly.
	\subsection*{Implementation Details}
	We have implemented the network using Tensorflow \cite{tensorflow}. Thus, the Fourier and inverse Fourier transform are layers provided by the Tensorflow framework. The parallel projector and back-projector as well as the fan-beam projector and back-projector are unmatched pairs and are implemented as custom ops in Tensorflow using Cuda kernels. For the back-propagation the respective operation is assigned to the layers for the gradient calculation.
	\subsection{Training Process}
	\subsection*{Training Data}
\begin{figure}		
		\centering
		\subfloat[Circle]{\includegraphics[width=0.17\textwidth]{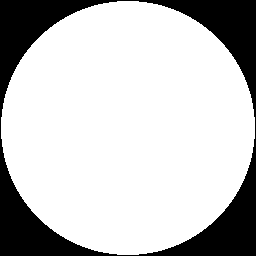}\label{circle_reco_gt}}
		\hspace{0.2cm}
		\subfloat[Ellipse]{\includegraphics[width=0.17\textwidth]{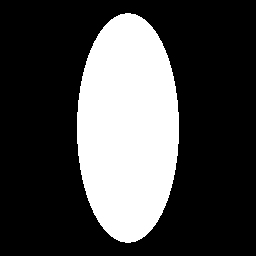}}
		\hspace{0.2cm}
		\subfloat[Ellipse-bar]{\includegraphics[width=0.17\textwidth]{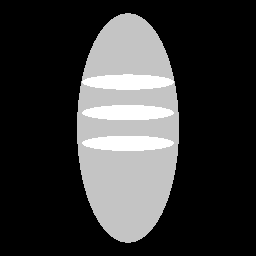}}
		\hspace{0.2cm}
		\subfloat[Bar]{\includegraphics[width=0.17\textwidth]{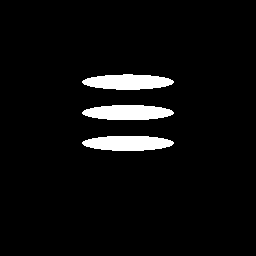}}
		\hspace{0.2cm}
		\subfloat[Noise]{\includegraphics[width=0.17\textwidth]{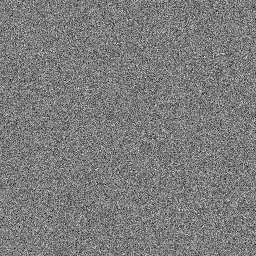}}
		\caption{Phantom types in training set.}
		\label{fig:phantoms}
	\end{figure}

	For the training we use numerical phantoms which bring their different characteristics into the training process (Fig.\ref{fig:phantoms}). The first type of phantoms are homogeneous objects that fill the field of view like ellipses and circles. The second type contain a homogeneous field of view filling ellipse with contains varying number of elongated ellipses (in the following called bars). The third type of phantoms uses only bars without the surrounding ellipsoid. As a last type we use phantoms which contain normal distributed noise.\\
	In the following list, the number test phantoms are listed:
	\begin{itemize}
		\item 1 Ellipse phantom
		\item 1 Circle phantom
		\item 8 Ellipse-bar phantoms (with increasing number of bars from 1 up to 8)
		\item 5 Bar phantoms (with increasing number of bars from 1 to 5
		\item 50 Noise phantoms (Normal distributed noise)
	\end{itemize}
	The parallel projections and respective label projections (fan-beam) are based on the following  geometry:
	\begin{itemize}
		\item Trajectory: [$0^\circ$, $25^\circ$, $45^\circ$, $65^\circ$, $90^\circ$]
		\item Source detector distance (SDD): $1200\ $mm
		\item Source isocenter distance (SID): $900\ $mm
		\item Parallel and fan-beam detector size: 512 pixel
		\item Reconstruction size: $256\times 256$
	\end{itemize}
	Thus, the training data set consists of partial parallel projections according to the method described in \cite{filter_learning} using the given angles in the trajectory for $\beta$. The respective label fan-beam projection is generated for each angle of the trajectory for each phantom. All projections are generated using the implemented projection layers.
	The performance of the network is validated using the Shepp-Logan phantom \cite{SheppLogan}.
	\subsection*{Training Setup}	
	The training process is divided into two steps. In the first step, the scaling layer $\bm S$ is trained while $\bm K$ remains fixed. After the training of the scaling layer $\bm S$ converges, the scaling factor is fixed and the training of the filter $\bm K$ is started. This separation is based on two thoughts. First, the scaling layer fixes an occurring problem due to the mix up of the different forward- and backprojection geometries and is not part of the unknown operator. The second point is that by dividing the learning process into two parts the learning rate for the scaling layer can be much higher and therefore speed up the whole training process. Furthermore the separation ensures that the calculated loss w.r.t. the label projection can express the deviation from the real fan-beam projection and is not distracted by a scaling factor due to the mixed forward- and backprojection. The filter $\bm K$ is initialized with the Ram-Lak filter \cite{RamLak}, which is an optimal discrete reconstruction filter for a complete data acquisition and therefore can be interpret as a strong pre-training of the network. We train on different sub-sampling factors, starting with full sampling and continuing by successively sub-sampling to 15, 7, 5 and 3 projections. This allows us to compare with the geometrical rebinning approach \cite{igic}.
	
	\subsection*{Projection-dependent Vs. Projection-independent}   
	To determine which type of filter performs the best, we performed all experiments on the different sub-sampling levels using both a projection-dependent and a projection-independent of version of $\bm K$.
	\subsection*{Regularization}
	To achieve smooth filter weights we use a Gaussian smoothing after each training epoch.
	
	\section{Results}
	The performance of the network is evaluated in three steps. First we analyze the performance for the different sub-sampling stages using the Shepp-Logan phantom and the fan-beam forward projection of the phantom as ground truth (GT). Afterwards, the results are compared with the geometrical approach with certain sub-sampling factors. To provide a better qualitative impression of the performance we subsequently present a comparison based on a 3D phantom using a stacked fan-beam approach. The network performance analysis is followed by a presentation of the learned filter types.    
	\subsection*{Network Performance}
\begin{figure}
		\subfloat{\includegraphics[width=\linewidth]{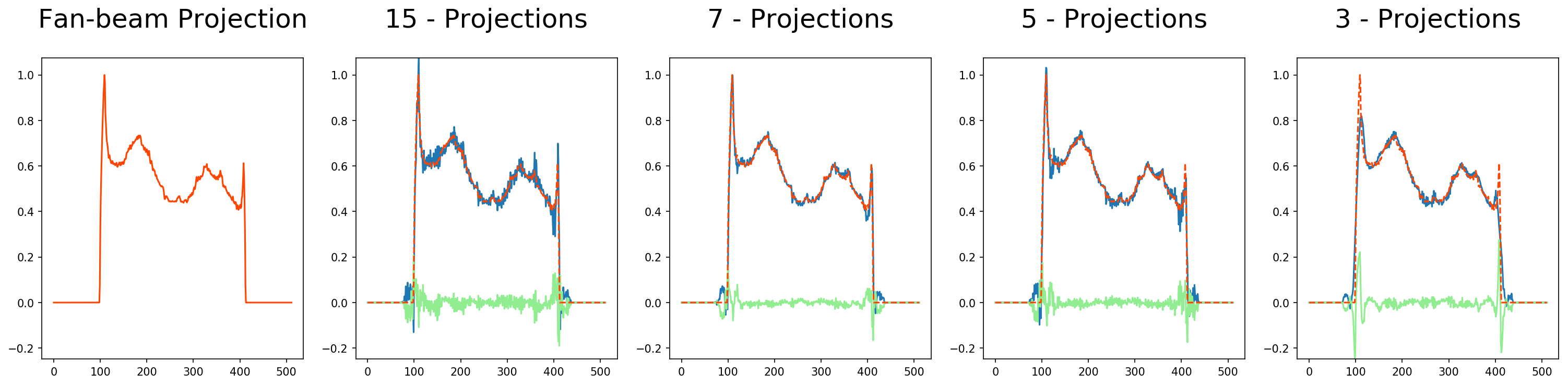}}
		\caption{Sub-sampling comparison of the projection-dependent filter with the GT projection of the Shepp-Logan phantom. The plot colors are red for the reference, blue for the respective line profile and green for the difference.}
		\label{fig:shift_variant_undersampling_shepp}
	\end{figure}

\begin{figure}
		\subfloat{\includegraphics[width=\linewidth]{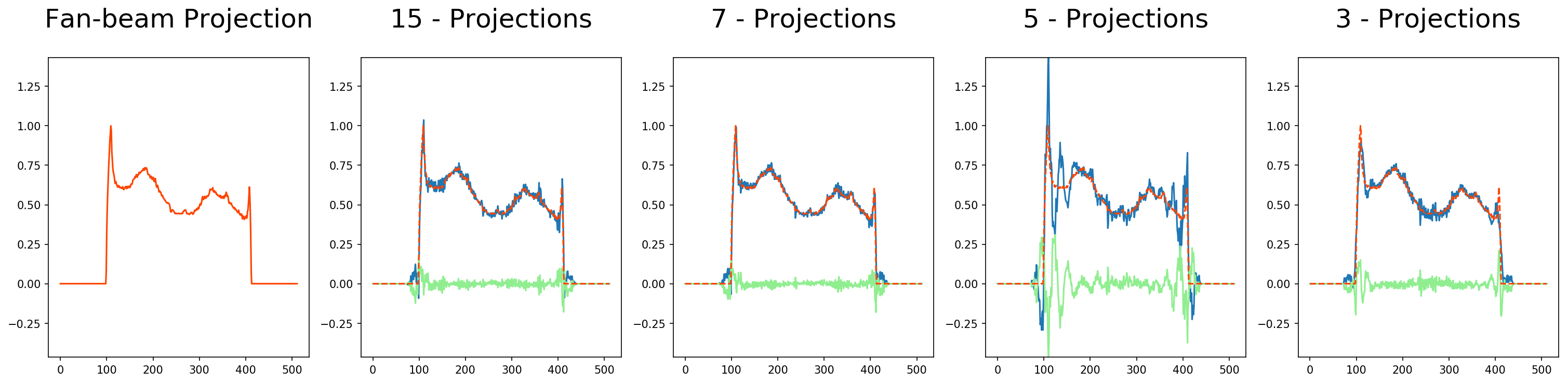}}
		\caption{Sub-sampling comparison of the projection-independent filter applied on the Shepp-Logan phantom. The plot colors are red for the reference, blue for the respective line profile and green for the difference.}
		\label{fig:shift_invariant_undersampling_shepp}
	\end{figure}

	In Fig. \ref{fig:shift_variant_undersampling_shepp} the rebinning performance of the learned network for the projection-dependent filter using the Shepp-Logan phantom with different sub-sampling factors is shown. All results show a similar shape as the line profile of the GT projection. The full sampling as well as the sub-sampling case using 15 projections show a noisy behavior. The noise is less for the sub-sampling cases using 7, 5 and 3 projections, respectively. For all versions, except for the case with 7 projections, the rebinned signal overshoots the GT signal at the edges of the object.
	For the projection-independent version of the filter (Fig. \ref{fig:shift_invariant_undersampling_shepp}) similar but strengthened behavior can be observed. For all four rebinning types the projection-independent counterpart is more noisy and overshoots or undershoots more extensively, especially for the rebinning with 5 projections.
	
\begin{figure}
		\subfloat{\includegraphics[width=\linewidth]{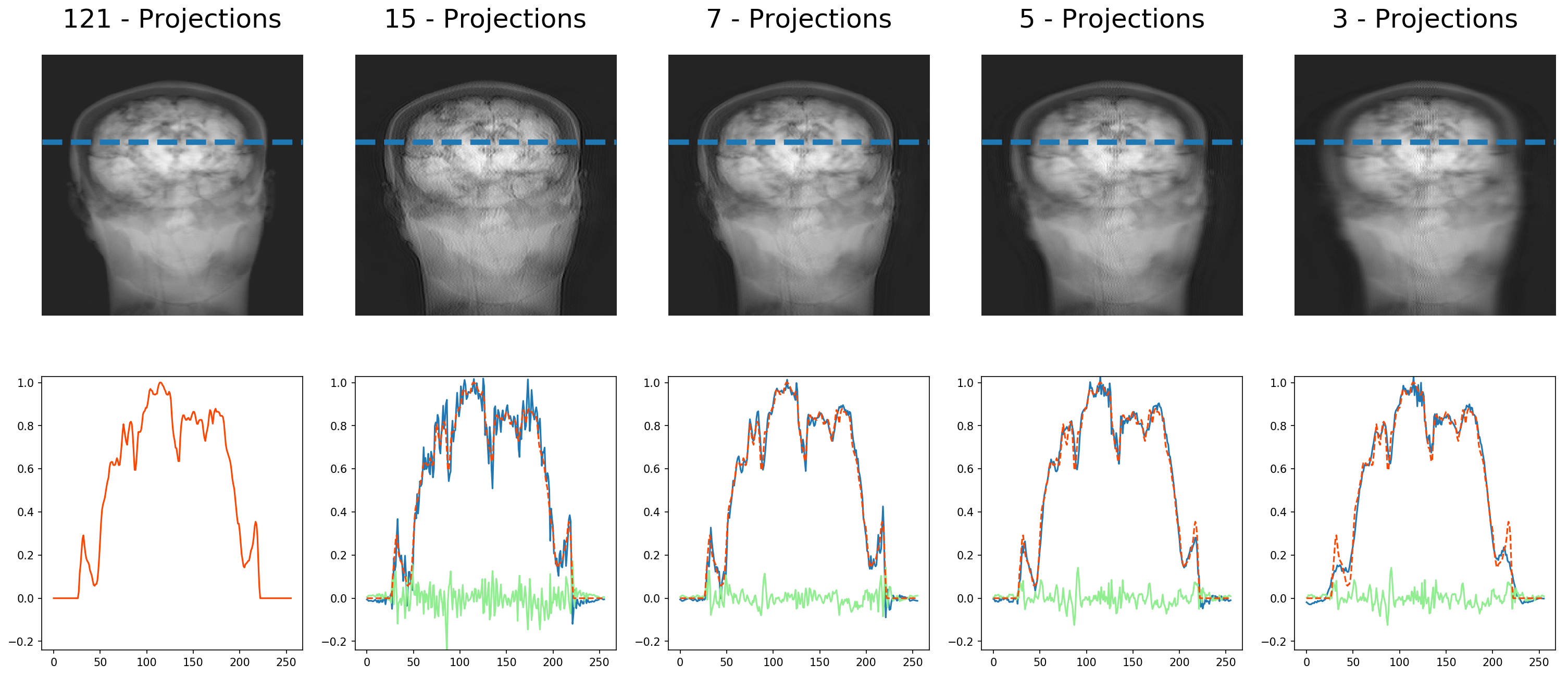}}
		\caption{Sub-sampling comparison of the projection-dependent filter applied on the MR head phantom. The plot colors are red for the reference, blue for the respective line profile and green for the difference.}
		\label{fig:shift_variant_undersampling_real}
	\end{figure}

	However, the noisiness of the 1D plots is misleading as the visual impression of the rebinned MR-projections from the head phantom show in Fig. \ref{fig:shift_variant_undersampling_real}. Even though the noisy behavior of the previous evaluation can be observed in the line profiles of the different sub-sampling methods, the noise level is not the main factor of the observed image impression. The experiment with 15 projections gives a sharp visual impression of the object, although it suffers from the strongest noise. The line profiles of the network trained with 5 and 3 projections show a reduced noise level compared to the network using 15 projections, but high-frequency artifacts and blurriness towards the edges of the image can be observed in the image. 
	
\begin{figure}
		\subfloat{\includegraphics[width=\linewidth]{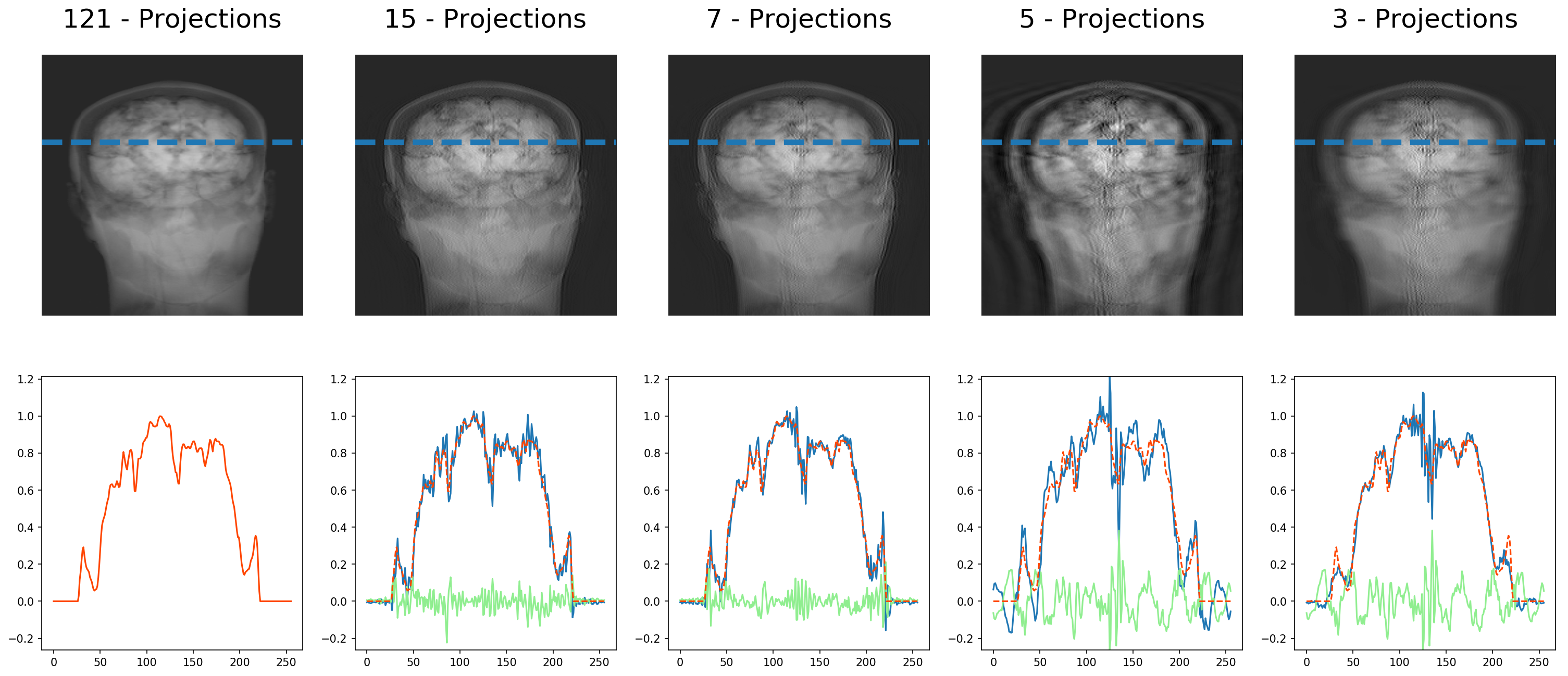}}
		\caption{Sub-sampling comparison projection-independent filter, real data. The plot colors are red for the reference, blue for the respective line profile and green for the difference.}
		\label{fig:shift_invariant_undersampling_real}
	\end{figure}

	For the projection-independent filter results, a similar but strengthened behavior can be observed in Fig. \ref{fig:shift_invariant_undersampling_real}. The filter for 15 projections provides a similar visual impression as the projection-dependent counterpart. The strength of the noise is stronger for the filter with 7, 5, and 3 projections than their respective projection-dependent counterpart. The high frequency artifacts are much stronger for the case with 5 and 3 projections. 
	
\begin{figure}
		\centering
		\subfloat{\includegraphics[width=0.8\linewidth]{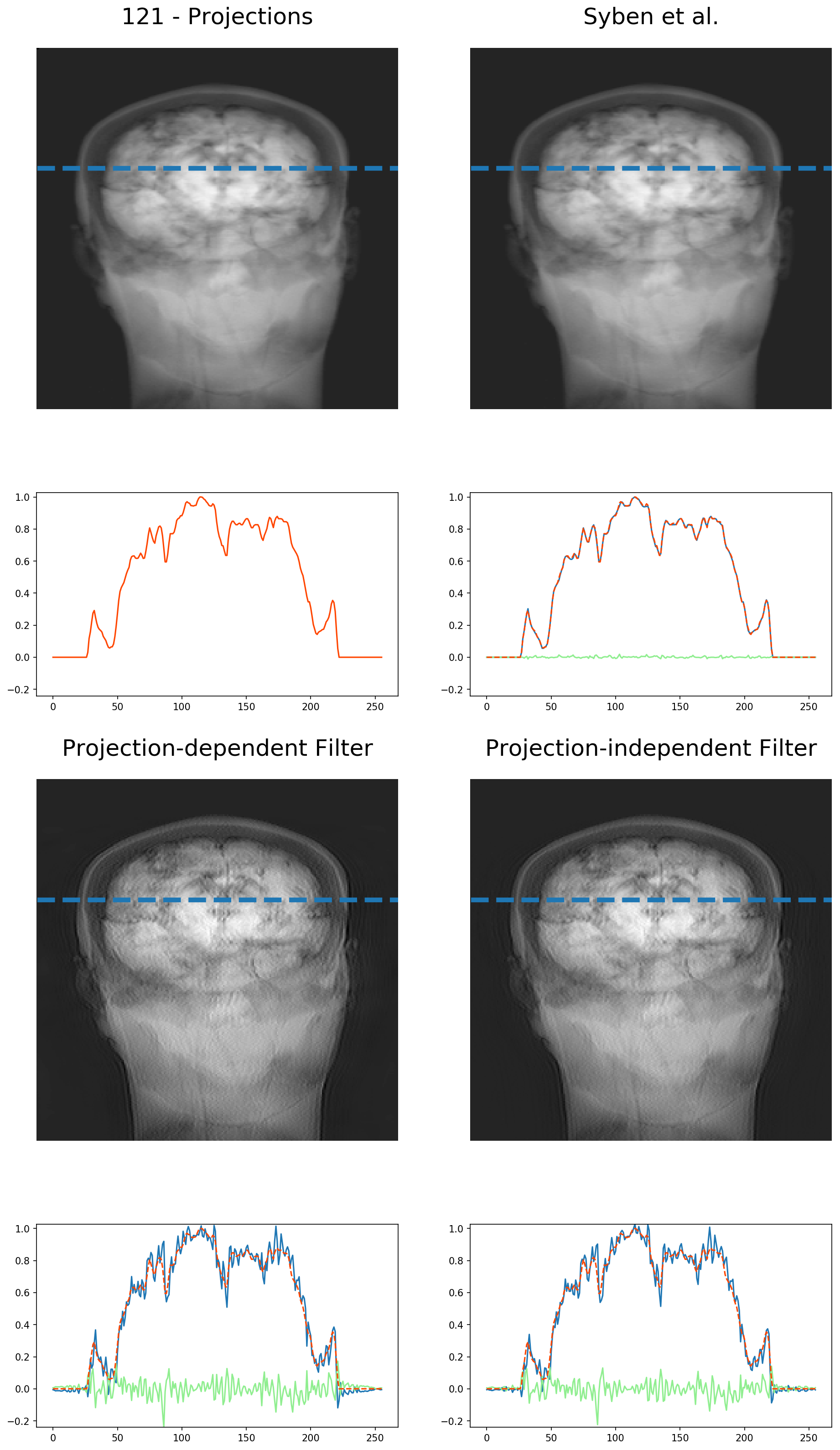}}
		\caption{Comparison of the geometrical rebinning methods and filter type using 15 projections out of the acquired 121 MR. The reference image with 121 projections is created by the geometrical rebinning method. The plot colors are red for the reference, blue for the respective line profile and green for the difference.}
		\label{fig:comparison_real}
	\end{figure}

	In Fig. \ref{fig:comparison_real} both filter types, projection-independent and projection-dependent are compared to the performance of the geometrical rebinning \cite{igic}. For the experiment 15 out of the acquired 121 projections of the head phantom are used. Both filters provide a sharper image impression compared to the reference method. In comparison with the geometrical rebinning method the results of both filters show high frequency artifacts at the edges of the phantom, which can be also seen in the line profiles.
	
	\subsection*{Filter Appearance}
\begin{figure}
		\subfloat{\includegraphics[width=\linewidth]{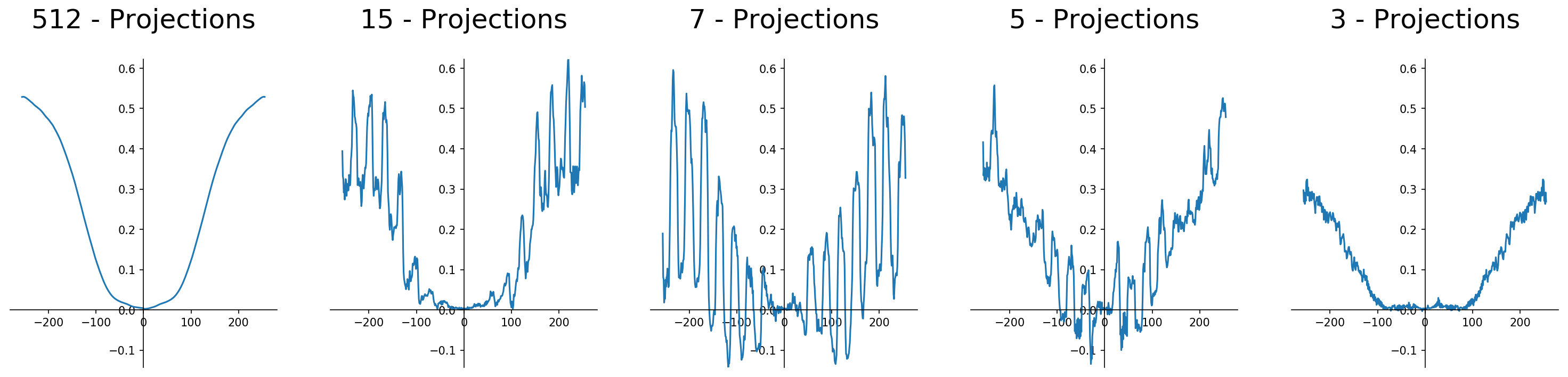}}
		\caption{Learned projection-independent filter in frequency domain for different sub-sampling factors.}
		\label{fig:shift_invariant_filter}
	\end{figure}

\begin{figure}
		\subfloat{\includegraphics[width=\linewidth]{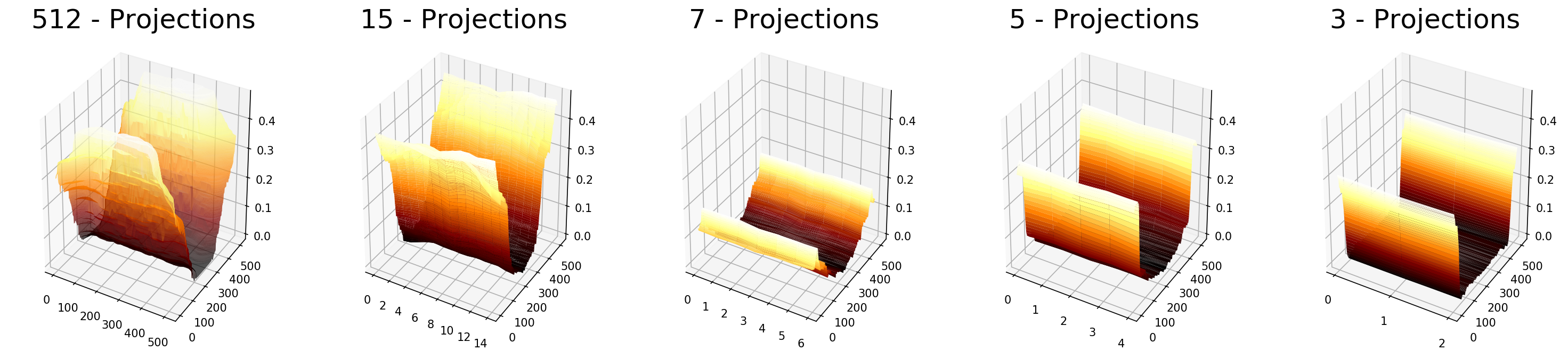}}
		\caption{Learned projection-dependent filter in frequency domain for different sub-sampling factors.}
		\label{fig:shift_variant_filter}
	\end{figure}

	In Fig. \ref{fig:shift_invariant_filter} the different learned projection-independent filters are shown. The filter using 512 projections is very smooth, while the filters with 15, 7, and 5 projections show high frequency components with a large amplitude. The filter for 3 projections has a high frequency component too, but with a much smaller amplitude. Furthermore, the amplitude of the filter is decreased compared to the initialization and the other filters.
	The learned projection-dependent filters are shown in Fig. \ref{fig:shift_variant_filter}. The filter for 512 projections shows in the middle a shape like the projection-independent counterpart, but drops off at the edges. While this is also true for the filter for 15 projections, the filter for 7, 5 and 3 projections are converging towards a U-shape.
	\section{Discussion}
	The results of the 1D fan-beam projections prove that our proposed analytical description of the rebinning process can be carried out learning the unknown operators in the problem description. The results of the MR head phantom provide a sharper visual impression than the rebinning method proposed Syben et al., although the noise level in the line profiles is much higher. The blurry visual impression of their approach is linked to the necessary interpolation in their method. Especially for image-guided interventions the sharpness is important to provide a clear impression of the vessels and interventional devices. Although the line profile for geometric rebinning overlaps very well with the projection reference, it must be taken into account that the reference case was already rebinned with this method itself based on all existing MR projections and is, therefore, already smoothed. Overall, using the learned filter based on 15 projections provides the best visual impression, while the amount of necessary MR projections is small. These observation are confirmed by the analysis published in \cite{igic}. In general, additional reduction of the number of projections is desirable, which could be achieved by further improving the filter learning process, e.g. by linking it directly to the k-space acquisition scheme.
	\\
	The results of the 1D projections as well as the stacked fan-beam experiment encourage a detailed discussion of the filter, its shape and the applied regularization. 
	The smoothing after each epoch leads to smooth filter weights for the projection-dependent case and also for the projection-independent filter for the full sampling. However, the smoothing does not enforce a smooth filter function for the projection-independent sub-sampling filters. Especially the 7 and 5 projection case show strong changes in amplitude. In the course of the experiments, we investigated different regularization terms, like the $\ell$2-norm on the filter weights or the $\ell$1-norm of the first derivative of the filter. However, regularization with the aformentioned methods performed not as well as expected. Despite thorough analysis of other regularization terms and corresponding weighting factors, the Gaussian smoothing lead to a more stable learning process and better results. However, a more profound method to achieve smooth filter weights is desirable. For this we started to look closer into regularizing the filter using the Lipschitz continuity. Certainly a more consistent regularization especially for the projection-independent filter has to be found. Such a regularization could open the opportunity to reduce the number of used projections in the rebinning process while preserving the sharp visual impression. Furthermore, introducing a symmetry constraint for the filter could improve the learning behavior and the outcome filter shape, while at the same time the number of parameters which have to be learned are reduced by a factor of 2.
	\\
	The results lead to several interesting questions which should be considered in further research. The impact of the number of used projections on the rebinning process as well as the covered frequency spectrum of the used phantoms on the filter shape are a promising line for subsequent work. The observed artifacts and the high frequency component in the projection-independent filter could be caused by insufficient coverage of the frequency space in the training process. Also the selection of the projections means a certain coverage of the wedge in the Fourier space as proposed by Syben et al. Furthermore the shape of the projection-dependent filter compared the 512 projection with the 3 projection filter version invites for further experiments. 
	The U-like shape of the projection-dependent filter in the 5 and 3 projection cases removes large amounts of low frequencies. With regard to MR acquisition, this could lead to a higher acquisition speed, as fewer frequencies have to be acquired in the K-space. Similar thoughts can be made with respect to the projection-independent filter with 7 projections. While it is more likely that the high change in amplitude is linked to the above discussion of the frequency spectrum and selected projections the question arises if a introduction of sparsity could lead to a sparse selection of frequencies.

	Note that this is not the only approach for fan-beam MR imaging. Wachowicz et al. \cite{beams_view} propose a method using additional non-linear gradient coils to directly acquire distorted images. Their approach is based on additional hardware, while we are demonstrate an acquisition approach which can be achieved without additional hardware.

	An overall interesting observation is the performance of the derived network topology. The results show that we can substitute the inverse bracket of right inverse of the system matrix by a filter in the frequency domain. The network topology to learn such a filter could be derived used the precision learning approach introduced in \cite{precision_learning}.
	
	\section{Conclusion}
	\label{sec:concl}
	We presented an alternative description of the rebinning process in terms of a projection-dependent or -independent filter. Based on the reconstruction problem and the problem description, we derived a network topology which allows to learn the unknown operators. Our proposed method provides a sharper image impression than the state-of-the-art method, since the necessary interpolation and thus smoothing steps can be avoided. Furthermore, the filter design is done entirely data-driven. The presented results encourage further investigation of the method. With deeper insight in the learning process, we assume that a further reduction of the necessary number of projections without losing the sharp image impression is possible. Additionally, as a next step the filter learning process may be extended to cone-beam projections. We hope that a better understanding of the filter will enable us to further reduce the number of data points to be recorded in k-space and, in the best case, to reduce them to points analytically determined by the filter. In the future, we want to combine our approach with MR acquisition trajectories specially adapted to our case. 
	
	Overall the results encourage to apply the proposed concept of learning unknown operators in domains where prior knowledge is available.
	
	\section*{Acknowledgement}
	This work has been supported by the project P3-Stroke, an EIT Health
	innovation project. EIT Health is supported by EIT, a body of the European
	Union.
	%
	%
	%
	\bibliographystyle{splncs04}
	\bibliography{0113}
	
\end{document}